\documentclass[journal]{IEEEtran}

\ifCLASSINFOpdf
\else
   \usepackage[dvips]{graphicx}
\fi
\usepackage{url}

\usepackage{graphicx}

\usepackage{booktabs}
\usepackage{multirow}
\usepackage{rotating}
\usepackage{mathrsfs,amssymb,amsmath, amsfonts,amsthm}

\begin{document}

\title{Extended Cross-Modality United Learning for Unsupervised Visible-Infrared Person Re-identification}

\author{Ruixing Wu, Yiming Yang, Jiakai He and Haifeng Hu, \IEEEmembership{Member, IEEE}
\thanks{Ruixing Wu, Jiakai He and Haifeng Hu are with the School of Electronics and Information Technology, Sun Yat-sen University, Guangzhou 510006, China (e-mail: wurx29@mail2.sysu.edu.cn; he.jk@qq.com; huhaif@mail.sysu.edu.cn).}
\thanks{Yiming Yang is with the School of Science and Engineering, The Chinese University of Hong Kong (Shenzhen), Shenzhen 518172, China (e-mail: yimingyang@link.cuhk.edu.cn).}
}

\markboth{Journal of \LaTeX\ Class Files, Vol. 14, No. 8, August 2015}
{Shell \MakeLowercase{\textit{et al.}}: Bare Demo of IEEEtran.cls for IEEE Journals}
\maketitle

\begin{abstract}
Unsupervised learning visible-infrared person re-identification (USL-VI-ReID) aims to learn modality-invariant features from unlabeled cross-modality datasets and reduce the inter-modality gap. However, the existing methods lack cross-modality clustering or excessively pursue cluster-level association, which makes it difficult to perform reliable modality-invariant features learning. To deal with this issue, we propose a Extended Cross-Modality United Learning (ECUL) framework, incorporating Extended Modality-Camera Clustering (EMCC) and Two-Step Memory Updating Strategy (TSMem) modules. Specifically, we design ECUL to naturally integrates intra-modality clustering, inter-modality clustering and inter-modality instance selection, establishing compact and accurate cross-modality associations while reducing the introduction of noisy labels. Moreover, EMCC captures and filters the neighborhood relationships by extending the encoding vector, which further promotes the learning of modality-invariant and camera-invariant knowledge in terms of clustering algorithm. Finally, TSMem provides accurate and generalized proxy points for contrastive learning by updating the memory in stages. Extensive experiments results on SYSU-MM01 and RegDB datasets demonstrate that the proposed ECUL shows promising performance and even outperforms certain supervised methods.
\end{abstract}


\IEEEpeerreviewmaketitle

\section{Introduction}
\IEEEPARstart{P}{erson} re-identification (ReID) is an important research direction in the field of computer vision, aiming to retrieve objects similar to a given surveillance pedestrian image from image or video sequences \cite{ye2021deep, yang2022augmented, yang2023towards, hou2024three, hao2021cross, ge2022structured, gwon2024balanced}. However, traditional single-modality ReID may not work effectively in night time or under poor lighting conditions. Thanks to recently deployed cameras that can automatically switch to far/near infrared mode at night \cite{yang2023dual, ye2021dynamic}, visible-infrared person re-identification (VI-ReID) is proposed to match pedestrian images of the same identity between the visible and infrared modalities \cite{ye2020dynamic, gao2021mso, wei2021syncretic, zhang2023diverse, kim2023partmix}. This technology is of great significance for application scenarios such as smart security and video surveillance \cite{qin2022deep, liang2021homogeneous}. However, annotations for cross-modal data sets require more resources than annotations for single-modal ReID data sets. 

The current VI-ReID methods, based on supervised learning of large amounts of manually labeled cross-modal data, have been able to provide reliable recognition results. This introduces a problem of learning modality-invariant knowledge from the unlabeled visible-infrared dataset, \textit{i.e.}, unsupervised learning visible-infrared person re-identification (USL-VI-ReID). In USL-VI-ReID, intra-modality clustering is a mainstream step. Some methods adopt cross-modality instance selection \cite{yang2022augmented, si2023diversity} or graph matching \cite{wu2023unsupervised, cheng2023efficient, yang2023robust} techniques for cross-modal association, which is difficult to handle cross-modal variation due to lacking compact mixed-modal training. Subsequently, many efforts employ inter-modality clustering \cite{pang2023cross, Yang2024DynamicMI} to establish the cross-modality relationships. However, the distance of the positive image pairs could be larger than that of the negative image pairs in USL-VI-ReID \cite{liang2021homogeneous}. These approaches crudely pursue identity consistency, which will introduce noise labels and confuse the inter-modality learning. In addition, the accuracy of the clustering results and the selection of clustering proxies also affect the ability of the model to distinguish hard samples. Lots of the existing methods \cite{yang2022augmented, yang2023towards, yang2023dual, wu2023unsupervised} directly use DBSCAN clustering algorithm and momentum update strategy, but ignore to improve them. Yin et al. \cite{yin2023real} proposed a real-time update strategy to solve the inconsistency between the clustering algorithm and the momentum memory update strategy. However, this approach simply replaces the memory entry with the current feature, which is prone to local bias and affects the generalization. 
\begin{figure*}
\centerline{\includegraphics[width=0.8\textwidth]{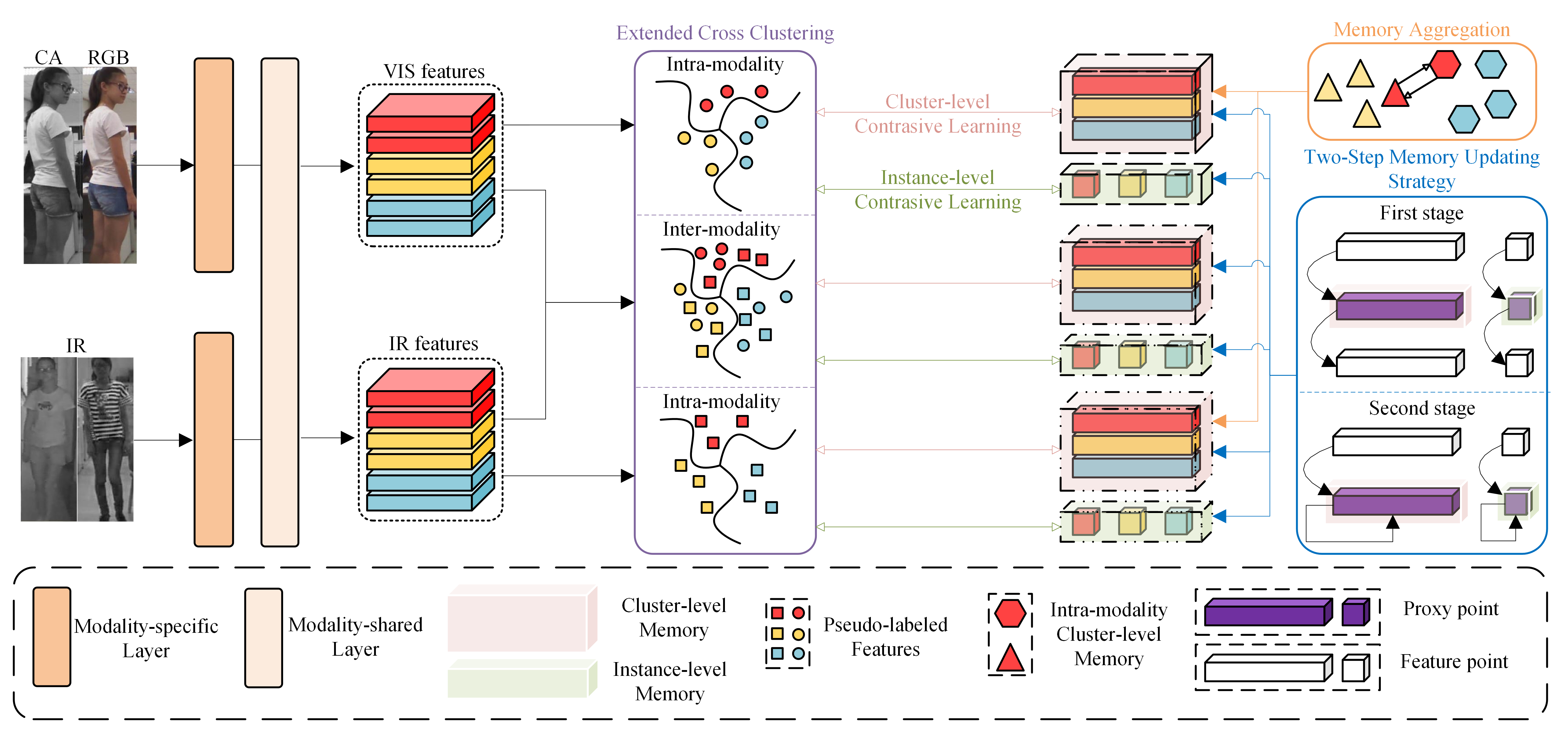}}
\caption{The structure of our ECUL framework for USL-VI-ReID. The ECUL framework is based on the cluster-level and instance-level contrastive losses and it integrates cross-modal clustering and inter-modality instance selection. EMCC and TSMem are included in the framework.}
\label{figure1}
\end{figure*}

To address the above challenges, we propose a comprehensive Extended Cross-Modality United Learning (ECUL) framework, incorporating Extended Modality-Camera Clustering (EMCC) and Two-Step Memory Updating Strategy (TSMem) modules.  Specifically, we design ECUL based on the cluster-level and instance-level contrastive losses to naturally integrates intra-modality clustering, inter-modality clustering and inter-modality instance selection. ECUL utilizes cross-modal clustering to cultivate modal invariant representations, while naturally combining improved cross-modality instance selection to reduce the introduction of noisy labels. Consequently, the combination make it cultivate modality-invariant representations and effectively improve the performance of the model. To further reduce the accumulation of unexpected noise labels, we deploy EMCC to improve the clustering algorithm. The EMCC can filter out the hard negative information by imposing additional restrictions, and then can obtain and fuse cleaner positive information from different modalities and cameras. Therefore, the cross-modality and cross-camera discrepancies can be mitigated. Finally, in order to maintain the accuracy of the proxy points stored in memory, we introduce TSMem to perform memory updates. In the early and late stages of training, we updated the memory with randomly sampled features and linear update strategies, respectively, thus balancing the differentiability and generalization of different memories. The main contributions can be summarized as follows:
\begin{itemize}
    \item We propose a comprehensive ECUL framework for USL-VI-ReID. This framework naturally fuses intra-modality clustering, inter-modality clustering and cross-modality instance selection to obtain clear and compact cross-modal learning, thus reliably bridging the cross-modality gaps under unsupervised condition.
    \item The EMCC improves the clustering algorithm by extending the encoding vector to fuse positive information and filter out hard negative information, thus reducing cross-modality and cross-camera discrepancies at the same time.
    \item The TSMem updates memory in two steps, providing accurate and generalized proxy points for contrastive learning.
    \item Extensive experimental results on SYSU-MM01 and RegDB datasets show that our method achieves competitive performance compared with the state of the arts.
\end{itemize}

\section{Methodology}

In this section, we will provide a detailed introduction to the ECUL framework, as depicted in Fig. \ref{figure1}. The ECUL framework integrates contrastive learning and memory aggregation. In addition, our ECUL also includes EMCC and TSMem, which will be elaborated in detail in Sections \ref{sec:II-B} and \ref{sec:II-C}, respectively.

\subsection{United Contrastive Learning and Memory Aggregation}
We first introduce the united contrastive learning and memory aggregation of our ECUL framework. Unlike the mainstream US-VI-ReID approach, ECUL naturally combines intra-modality clustering, inter-modality clustering and inter-modality instance selection. The combination can obtain compact mode-invariant associations while effectively avoiding confounding cross-modal learning. 

With the intra-modal pseudo labels and inter-modal pseudo labels, we divide the training architecture into intra-modality training and inter-modality training. In both sets of training, we compute the contrastive losses for visible modality, infrared modality and mixed modality by the following equations:
\begin{equation}
\mathcal{L}_{q} = - \log \frac{\exp(q \cdot \phi_{+} / \tau)}{\sum_{k=0}^{K} \exp(q \cdot \phi_{k} / \tau)}
\end{equation}
where $q$ is the query features. $\phi_+$ denotes the positive memory corresponding to the pseudo label of the query $q$ and $\tau$ is a temperature \cite{dai2022cluster}. Inspired by \cite{Yang2024DynamicMI}, we also deployed both instance-level and cluster-level memory initialization and contrastive losses during the training process.

In order to reduce the introduction of noise, we explore the correlation between the samples of two modalities. Firstly, we calculate the similarity of cross-modal instance pairs and select the memories to be aggregated according to the method of count priority selection \cite{yang2022augmented}. The selected memories are then aggregated by an improved memory aggregation, which can be denoted as:
\begin{equation}
    \phi_n^{i(t)} \leftarrow \alpha \phi_n^{i(t)} + (1 - \alpha) \phi_m^{v(t)}
\end{equation}
\begin{equation}
    \phi_m^{v(t)} \leftarrow \alpha \phi_m^{v(t)} + (1 - \alpha) \phi_n^{i(t)}
\end{equation}
where $\alpha$ denotes the momentum updating factor. $\phi_n^{i(t)}$ and $\phi_m^{v(t)}$ represent selected infrared and visible memories at epoch $t$. It is worth noting that our improved memory aggregation employ cross update between two modalities so that the model can learn better modality-invariant knowledge.

\subsection{Extended Modality-Camera Clustering}\label{sec:II-B}


In the clustering stage, the existing unsupervised ReID methods typically utilize the extracted features to compute the k-reciprocal encoding first. However, Cross-modality and cross-camera datasets may cause the distance within the same class larger than the distance between different classes, resulting in a large number of noisy identities associated during the clustering phase. In order to solve this problem, we introduce Extended Modality-Camera Clustering (EMCC) to improve the clustering algorithm. 

Specifically, each feature is initially regarded as a probe and subsequently computes the k-reciprocal encoding vector\cite{zhong2017re} in relation to other features with a proximity range $k_1$. Then we aggregate the distance coding information from the nearest neighbors to associate the hard samples. For intra-modality clustering, the process of extension can be formulated as follow:
\begin{equation}
    \tilde{\mathcal{V}}_i = \frac{1}{n^c} \sum_{j=1}^{n^c} \mathcal{V}_j
\end{equation}
\begin{equation}
    \mathcal{V}_i = \frac{1}{n_i} \sum_{j=1}^{k_3} \textbf{1}\left\{ l_j^{camera} = c \right\} \textbf{1}\left\{ s(j) \leq k_2 \right\} \mathcal{V}_j
\label{eq11}
\end{equation}
\begin{equation}
     s(i) = \sum_{j=1}^{i} \textbf{1}\left\{ l_j^{camera} = c \right\}
\label{eq12}
\end{equation}
where $n^c$ denotes the number of camera, $n_i$ indicates the number of samples that satisfy the indicated function, $l_j^{camera}$ represents the camera label, and $\textbf{1}\{\cdot\}$ is the indicator function. For inter-modality clustering, we need to multiply Eq. \ref{eq11} and Eq. \ref{eq12} by an indicator function $\textbf{1}\left\{ l_j^{modality} = d \right\}$ to identify different modalities. After we obtain the extended coding vector, the Jaccard distance between instances can be calculated, and then the pseudo-labels can be assigned.

\begin{figure}
\centerline{\includegraphics[width=0.75\columnwidth]{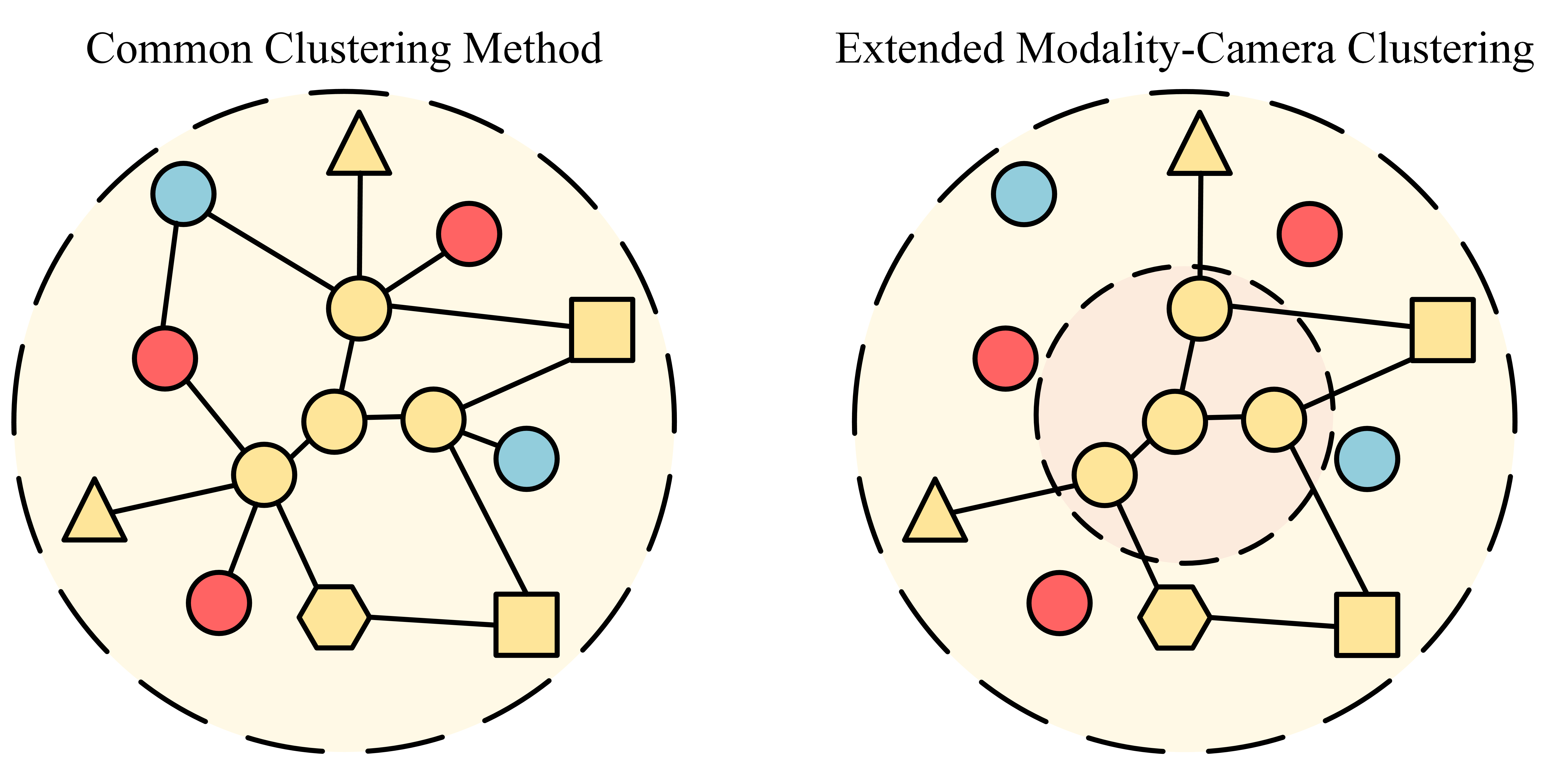}}
\caption{Illustration of the clustering results of common clustering method and our EMCC. Different colors represent different identities, while different shapes represent different modalities or cameras. In our EMCC, $k_2$ provides a small area range and $k_3$ provides a large area range, thus effectively filtering out hard negative information during the clustering phase.}
\label{figure2}
\end{figure}

With EMCC, we take into account information from different modalities and from different cameras by imposing $k_2$ and $k_3$ restrictions when extending the distance encoding. As shown in Fig. \ref{figure2}, $k_2$ provides a small area range to reduce the introduction of noise vectors, and $k_3$ provides a larger range to fuse positive information from different aspects. This approach improve the performance and robustness of the clustering algorithm, thereby reducing both the cross-modality and cross-camera discrepancies in clustering.

\subsection{Two-Step Memory Updating Strategy}\label{sec:II-C}


During training iterations, the memory is updated. Different from the common momentum update strategy \cite{he2020momentum} and real-time update strategy \cite{yin2023real}, our proposed TSMem is based on two parts: (1) update memory with current features. (2) update the memory with a linear update strategy.

In the early stages of training, the discriminative capability of the model is limited. Therefore, for the first 50 epochs, the features from the mini-batch are randomly sampled and used to directly replace the features from the memory. This method effectively retains the original feature distribution, which can not only cover more semantic information, but also keep the consistency of clustering and the memory updating to improve the model optimization accuracy \cite{yin2023real}. However, in the later stage of training, we should consider enhancing the generalization of the representative of the cluster center to avoid the center shift caused by too much attention to local information. Therefore we propose the linear update strategy. In the last 50 epochs of training, the linear update strategy requires the current memory to gradually increase the proportion of fused previous memories to ensure the freshness and generalization of the memory. The whole strategy of memory update can be expressed as follow.
\begin{equation}
    \phi_k^{(n)} \leftarrow q, e<50
\end{equation}
\begin{equation}
    \phi_k^{(n)} \leftarrow g(e) \phi_k^{(n-1)} + (1 - g(e)) q, e\ge50
\end{equation}
\begin{equation}
    g(e)=\frac{e-50}{e_t}+b
\end{equation}
where $q$ denotes the query features sampled from the current mini-batch. $n$ is the training iteration number. $e$ and $e_t$ denote the numbers of current and total epochs, respectively.

TSMem is applied to both instance-level memory update and cluster-level memory update, which provides accurate and generalized proxy points for contrastive learning.

\section{Experiments}
\subsection{Datasets and Settings}
\subsubsection{Dataset}
{The proposed ECUL framework is evaluated on two public US-VI-ReID benchmarks, \textit{i.e.}, SYSU-MM01 \cite{wu2017rgb} and RegDB \cite{nguyen2017person}. SYSU-MM01 is a large-scale image dataset collected by four RGB cameras and two infrared cameras. The dataset contains 287,628 visible and 15,792 infrared images of 491 identities. RegDB has a total of 4120 RGB images and 4120 infrared images. It contains 412 pedestrians with different identities, each corresponding to 10 RGB images and 10 infrared images.}

\subsubsection{Evaluation Metrics}
{In the experiments, Cumulated Matching Characteristics (CMC), Mean Average Precision (mAP) and 
and mean inverse negative penalty (mINP) \cite{ye2021deep} are employed as the evaluation metrics to compare our ECUL with other existing methods.}

\subsubsection{Implementation Details}
{Our ECUL framework is implemented on PyTorch and trained with 2 TITAN Xp GPU. We adopt the feature extractor in AGW \cite{ye2021deep}, which is based on ResNet-50 \cite{he2016deep} pre-trained by ImageNet. We also employ Channel exchangeable Augmentation (CA) \cite{ye2021channel}, which can effectively narrow the modal gap between infrared and visible modalities \cite{yang2022augmented, wu2023unsupervised}. The pseudo labels are assigned by DBSCAN \cite{ester1996density} at the beginning of each training epoch. For each batch, we select 8 individuals and 16 instances for each individual in the training phase. The learning rate is set to 0.00035 and it is decayed by 0.1 every 20 epochs. The $\alpha$ is set to 0.2. The initial values and adjustments of $eps$, $k_1$, $k_2$ and $k_3$ follow the approach of Dynamic Neighborhood Clustering (DNC)\cite{Yang2024DynamicMI}. The specific settings can be found in our code.

\subsection{Comparison With the State-of-the-Art Methods}
As shown in Table \ref{table1}, our ECUL is compared with the state-of-art VI-ReID methods on SYSU-MM01 and RegDB.
\subsubsection{Comparison With Unsupervised Methods}
In general, ECUL outperforms the current advanced unsupervised methods. Our method achieved a $66.01\%$ rank-1 accuracy on SYSU-MM01 (all search), which is $2.83\%$ and $2.5\%$ higher than DCCL and GUR respectively. In particular, the rank-1 accuracy on RegDB (visible to infrared) of ECUL is much higher than DCCL and GUR, achieving a remarkable accuracy of $87.30\%$.

\begin{table}
\centering
\caption{Comparisons Of Our Method With The State-of-the-Art Methods On SYSU-MM01 And RegDB}
\label{table1}
\renewcommand{\arraystretch}{1.1}
\scalebox{0.67}{%
\setlength{\tabcolsep}{3pt}
\begin{tabular}{p{5pt}|p{45pt}|p{17pt}|p{17pt}|p{17pt}|p{17pt}|p{17pt}|p{17pt}|p{17pt}|p{17pt}|p{17pt}|p{17pt}|p{17pt}|p{17pt}}
\hline
&& \multicolumn{6}{|c|}{SYSU-MM01}& \multicolumn{6}{|c}{RegDB}\\
\cline{3-14}
& & \multicolumn{3}{|c}{All-search}&  \multicolumn{3}{|c|}{Indoor-search}& \multicolumn{3}{|c}{Visible to Infrared}& \multicolumn{3}{|c}{Infrared to Visible}\\
&Method & \multicolumn{1}{|c}{r1} & \multicolumn{1}{c}{mAP} & \multicolumn{1}{c}{mINP} & \multicolumn{1}{|c}{r1} & \multicolumn{1}{c}{mAP} & \multicolumn{1}{c}{mINP} & \multicolumn{1}{|c}{r1} & \multicolumn{1}{c}{mAP} & \multicolumn{1}{c}{mINP} & \multicolumn{1}{|c}{r1} & \multicolumn{1}{c}{mAP} & \multicolumn{1}{c}{mINP}\\
\hline

&AGW\cite{ye2021deep}
        & \multicolumn{1}{c}{47.50} & \multicolumn{1}{c}{47.65} & \multicolumn{1}{c|}{35.30} 
        & \multicolumn{1}{c}{54.17} & \multicolumn{1}{c}{62.97} & \multicolumn{1}{c|}{59.23} 
        & \multicolumn{1}{c}{70.05} & \multicolumn{1}{c}{66.37} & \multicolumn{1}{c|}{50.19} 
        & \multicolumn{1}{c}{70.49} & \multicolumn{1}{c}{65.90} & \multicolumn{1}{c}{51.24}\\
&DDAG\cite{ye2020dynamic}   
        & \multicolumn{1}{c}{51.24} & \multicolumn{1}{c}{53.02} & \multicolumn{1}{c|}{39.62} 
        & \multicolumn{1}{c}{61.02} & \multicolumn{1}{c}{61.02} & \multicolumn{1}{c|}{62.61} 
        & \multicolumn{1}{c}{69.34} & \multicolumn{1}{c}{63.46} & \multicolumn{1}{c|}{49.24} 
        & \multicolumn{1}{c}{68.06} & \multicolumn{1}{c}{61.80} & \multicolumn{1}{c}{48.62}\\
&MSO\cite{gao2021mso}    
        & \multicolumn{1}{c}{58.70} & \multicolumn{1}{c}{56.42} & \multicolumn{1}{c|}{-} 
        & \multicolumn{1}{c}{63.09} & \multicolumn{1}{c}{70.31} & \multicolumn{1}{c|}{-} 
        & \multicolumn{1}{c}{73.6} & \multicolumn{1}{c}{66.9} & \multicolumn{1}{c|}{-}
        & \multicolumn{1}{c}{74.6} & \multicolumn{1}{c}{67.5} & \multicolumn{1}{c}{-}\\
&MCLNet\cite{hao2021cross}
        & \multicolumn{1}{c}{65.40} & \multicolumn{1}{c}{61.98} & \multicolumn{1}{c|}{47.39} 
        & \multicolumn{1}{c}{72.56} & \multicolumn{1}{c}{76.58} & \multicolumn{1}{c|}{72.10} 
        & \multicolumn{1}{c}{80.31} & \multicolumn{1}{c}{73.07} & \multicolumn{1}{c|}{57.39} 
        & \multicolumn{1}{c}{75.93} & \multicolumn{1}{c}{69.49} & \multicolumn{1}{c}{52.63}\\
\turnbox{90}{Supervised}&SMCL\cite{wei2021syncretic}
        & \multicolumn{1}{c}{67.39} & \multicolumn{1}{c}{61.78} & \multicolumn{1}{c|}{-} 
        & \multicolumn{1}{c}{68.84} & \multicolumn{1}{c}{75.56} & \multicolumn{1}{c|}{-} 
        & \multicolumn{1}{c}{83.93} & \multicolumn{1}{c}{79.83} & \multicolumn{1}{c|}{-} 
        & \multicolumn{1}{c}{83.05} & \multicolumn{1}{c}{78.57} & \multicolumn{1}{c}{-}\\
&DEEN\cite{zhang2023diverse}   
        & \multicolumn{1}{c}{74.7} & \multicolumn{1}{c}{71.8} & \multicolumn{1}{c|}{-} 
        & \multicolumn{1}{c}{80.3} & \multicolumn{1}{c}{83.3} & \multicolumn{1}{c|}{-} 
        & \multicolumn{1}{c}{91.1} & \multicolumn{1}{c}{85.1} & \multicolumn{1}{c|}{-} 
        & \multicolumn{1}{c}{89.5} & \multicolumn{1}{c}{83.4} & \multicolumn{1}{c}{-}\\
&SGIEL\cite{kim2023partmix}  
        & \multicolumn{1}{c}{77.12} & \multicolumn{1}{c}{72.33} & \multicolumn{1}{c|}{-} 
        & \multicolumn{1}{c}{82.07} & \multicolumn{1}{c}{82.95} & \multicolumn{1}{c|}{-} 
        & \multicolumn{1}{c}{95.35} & \multicolumn{1}{c}{89.98} & \multicolumn{1}{c|}{-} 
        & \multicolumn{1}{c}{97.57} & \multicolumn{1}{c}{91.41} & \multicolumn{1}{c}{-}\\
\hline
&H2H\cite{liang2021homogeneous}    
        & \multicolumn{1}{c}{30.15} & \multicolumn{1}{c}{29.40} & \multicolumn{1}{c|}{-}     
        & \multicolumn{1}{c}{-}     & \multicolumn{1}{c}{-}     & \multicolumn{1}{c|}{-} 
        & \multicolumn{1}{c}{23.81} & \multicolumn{1}{c}{18.87} & \multicolumn{1}{c|}{-} 
        & \multicolumn{1}{c}{-} & \multicolumn{1}{c}{-} & \multicolumn{1}{c}{-}\\
&ADCA\cite{yang2022augmented}   
        & \multicolumn{1}{c}{45.51} & \multicolumn{1}{c}{42.73} & \multicolumn{1}{c|}{28.29} 
        & \multicolumn{1}{c}{50.60} & \multicolumn{1}{c}{59.11} & \multicolumn{1}{c|}{55.17} 
        & \multicolumn{1}{c}{67.20} & \multicolumn{1}{c}{64.05} & \multicolumn{1}{c|}{52.67} 
        & \multicolumn{1}{c}{68.48} & \multicolumn{1}{c}{63.81} & \multicolumn{1}{c}{49.62}\\
&MBCCM\cite{cheng2023efficient}
        & \multicolumn{1}{c}{53.14} & \multicolumn{1}{c}{48.16} & \multicolumn{1}{c|}{32.41} 
        & \multicolumn{1}{c}{55.21} & \multicolumn{1}{c}{61.98} & \multicolumn{1}{c|}{57.13} 
        & \multicolumn{1}{c}{83.79} & \multicolumn{1}{c}{77.87} & \multicolumn{1}{c|}{65.04} 
        & \multicolumn{1}{c}{82.82} & \multicolumn{1}{c}{76.74} & \multicolumn{1}{c}{61.73}\\
&PGM\cite{wu2023unsupervised}
        & \multicolumn{1}{c}{57.27} & \multicolumn{1}{c}{51.78} & \multicolumn{1}{c|}{34.96} 
        & \multicolumn{1}{c}{56.23} & \multicolumn{1}{c}{62.74} & \multicolumn{1}{c|}{58.13} 
        & \multicolumn{1}{c}{69.48} & \multicolumn{1}{c}{65.41} & \multicolumn{1}{c|}{-} 
        & \multicolumn{1}{c}{69.85} & \multicolumn{1}{c}{65.17} & \multicolumn{1}{c}{-}\\
\turnbox{90}{Unsupervised}&CHCR\cite{pang2023cross}
        & \multicolumn{1}{c}{59.47} & \multicolumn{1}{c}{59.14} & \multicolumn{1}{c|}{-} 
        & \multicolumn{1}{c}{-} & \multicolumn{1}{c}{-} & \multicolumn{1}{c|}{-} 
        & \multicolumn{1}{c}{69.31} & \multicolumn{1}{c}{64.74} & \multicolumn{1}{c|}{-} 
        & \multicolumn{1}{c}{69.96} & \multicolumn{1}{c}{65.87} & \multicolumn{1}{c}{-}\\
&DCCL\cite{yang2023dual}
        & \multicolumn{1}{c}{63.18} & \multicolumn{1}{c}{58.62} & \multicolumn{1}{c|}{42.99} 
        & \multicolumn{1}{c}{66.67} & \multicolumn{1}{c}{71.82} & \multicolumn{1}{c|}{67.46} 
        & \multicolumn{1}{c}{78.28} & \multicolumn{1}{c}{71.98} & \multicolumn{1}{c|}{58.79} 
        & \multicolumn{1}{c}{78.28} & \multicolumn{1}{c}{71.30} & \multicolumn{1}{c}{55.23}\\
&GUR\cite{yang2023towards}
        & \multicolumn{1}{c}{63.51} & \multicolumn{1}{c}{61.63} & \multicolumn{1}{c|}{47.93} 
        & \multicolumn{1}{c}{71.11} & \multicolumn{1}{c}{76.23} & \multicolumn{1}{c|}{72.57} 
        & \multicolumn{1}{c}{73.91} & \multicolumn{1}{c}{70.23} & \multicolumn{1}{c|}{58.88} 
        & \multicolumn{1}{c}{75.00} & \multicolumn{1}{c}{69.94} & \multicolumn{1}{c}{56.21}\\
\hline
&ECUL   & \multicolumn{1}{c}{\textbf{66.01}} & \multicolumn{1}{c}{\textbf{62.80}} & \multicolumn{1}{c|}{\textbf{48.47}} 
        & \multicolumn{1}{c}{\textbf{72.29}} & \multicolumn{1}{c}{\textbf{77.35}} & \multicolumn{1}{c|}{\textbf{73.51}} 
        & \multicolumn{1}{c}{\textbf{87.30}} & \multicolumn{1}{c}{\textbf{80.09}} & \multicolumn{1}{c|}{\textbf{66.11}} 
        & \multicolumn{1}{c}{\textbf{86.50}} & \multicolumn{1}{c}{\textbf{78.96}} & \multicolumn{1}{c}{\textbf{63.01}}\\
\hline
\multicolumn{14}{p{315pt}}{Rank at r accuracy(\%), mAP(\%) and mINP(\%) are reported. The best results are in bold.}\\
\end{tabular}
}
\label{table1}
\end{table}

\subsubsection{Comparison With Supervised Methods}
We also compare the proposed ECUL with some prevailing supervised methods. Table \ref{table1} can clearly indicate that the performance of our method exceeds several supervised methods, such as AGW \cite{ye2021deep}, MSO \cite{gao2021mso} and DDAG \cite{ye2020dynamic}. Experimental results similar to MCLNet \cite{hao2021cross} and SMCL \cite{wei2021syncretic} also show the competitive performance of our method. Compared with supervised methods, our method can eliminate the dependence on annotations and significantly reduce the cost.

\subsection{Ablation Study}
In this subsection, we conduct ablation experiments on the SYSU-MM01 dataset as a way to determine the contribution of different components in the proposed ECUL to the overall performance. The results are presented in Table \ref{table2}, where we employ Dual-Contrastive Learning (DCL) \cite{yang2022augmented} as our baseline. $\mathcal{L}_{I}$ is instance-level contrastive loss, and $\mathcal{L}^{m}$ is contrastive loss of inter-modality training. CMA and IMA represent Cross-modality Memory Aggregation \cite{yang2022augmented} and our improved memory aggregation, respectively.
\subsubsection{Effectiveness of United Contrastive Learning and Memory Aggregation}
Compared to the baseline, the performance is improved by $15.83\%$ and $4.99\%$ of rank-1 on all-search and indoor-search by adding $\mathcal{L}_{I}$ and $\mathcal{L}^{m}$. Moreover, there is an additional performance improvement of $10.78\%$ and $10.14\%$ with the incorporation of IMA. In addition, indices 3 and 4 denote that our IMA is more effective than the CMA. These results indicate that combining the above losses and memory aggregation together helps to extract valid identity information more accurately and adapt to modality differences.

\begin{table}
\centering
\caption{Ablation Studies For Objective Functions On SYSU-MM01}
\label{table2}
\renewcommand{\arraystretch}{1.1}
\scalebox{0.65}{%
\setlength{\tabcolsep}{4pt}
\begin{tabular}{p{10pt}|p{15pt}p{15pt}p{15pt}p{15pt}p{15pt}p{15pt}|p{15pt}|p{15pt}|p{15pt}|p{15pt}|p{15pt}|p{15pt}}
\hline
& \multicolumn{6}{|c}{Components}& \multicolumn{3}{|c}{All Search}& \multicolumn{3}{|c}{Indoor Search}\\
\hline
\multicolumn{1}{c}{Index} & \multicolumn{1}{|c}{Baseline} & \multicolumn{1}{c}{$\mathcal{L}_{I}+\mathcal{L}^{m}$} & \multicolumn{1}{c}{CMA} & \multicolumn{1}{c}{IMA} & \multicolumn{1}{c}{EMCC} & \multicolumn{1}{c}{TSMem} & \multicolumn{1}{|c}{r1} & \multicolumn{1}{c}{mAP} & \multicolumn{1}{c}{mINP} & \multicolumn{1}{|c}{r1} & \multicolumn{1}{c}{mAP} & \multicolumn{1}{c}{mINP}\\
\hline

\multicolumn{1}{c|}{1}    & \multicolumn{1}{c}{\checkmark} &  &  &  &  &
        & \multicolumn{1}{c}{32.60} & \multicolumn{1}{c}{32.90} & \multicolumn{1}{c|}{21.20} 
        & \multicolumn{1}{c}{37.54} & \multicolumn{1}{c}{47.01} & \multicolumn{1}{c}{43.39}\\
\multicolumn{1}{c|}{2}   & \multicolumn{1}{c}{\checkmark} & \multicolumn{1}{c}{\checkmark} 
        &  &  &  &
        & \multicolumn{1}{c}{48.43} & \multicolumn{1}{c}{47.16} & \multicolumn{1}{c|}{33.92} 
        & \multicolumn{1}{c}{52.53} & \multicolumn{1}{c}{60.76} & \multicolumn{1}{c}{56.82}\\
\multicolumn{1}{c|}{3}    & \multicolumn{1}{c}{\checkmark} & \multicolumn{1}{c}{\checkmark} 
        & \multicolumn{1}{c}{\checkmark} & & & 
        & \multicolumn{1}{c}{56.00} & \multicolumn{1}{c}{55.23} & \multicolumn{1}{c|}{41.96}
        & \multicolumn{1}{c}{61.95} & \multicolumn{1}{c}{69.10} & \multicolumn{1}{c}{65.06}\\
\multicolumn{1}{c|}{4}  & \multicolumn{1}{c}{\checkmark} & \multicolumn{1}{c}{\checkmark} &
        & \multicolumn{1}{c}{\checkmark} &  & 
        & \multicolumn{1}{c}{59.21} & \multicolumn{1}{c}{56.26} & \multicolumn{1}{c|}{41.69} 
        & \multicolumn{1}{c}{62.67} & \multicolumn{1}{c}{69.75} & \multicolumn{1}{c}{65.66}\\
\multicolumn{1}{c|}{5}  & \multicolumn{1}{c}{\checkmark} & \multicolumn{1}{c}{\checkmark} & 
        & \multicolumn{1}{c}{\checkmark} & \multicolumn{1}{c}{\checkmark} & 
        & \multicolumn{1}{c}{63.86} & \multicolumn{1}{c}{61.22} & \multicolumn{1}{c|}{47.79} 
        & \multicolumn{1}{c}{70.34} & \multicolumn{1}{c}{76.02} & \multicolumn{1}{c}{72.13}\\        
\hline
\multicolumn{1}{c|}{6}    & \multicolumn{1}{c}{\checkmark} & \multicolumn{1}{c}{\checkmark}  & 
        & \multicolumn{1}{c}{\checkmark} & \multicolumn{1}{c}{\checkmark} & \multicolumn{1}{c|}{\checkmark} 
        & \multicolumn{1}{c}{\textbf{66.01}} & \multicolumn{1}{c}{\textbf{62.80}} & \multicolumn{1}{c|}{\textbf{48.47}} 
        & \multicolumn{1}{c}{\textbf{72.29}} & \multicolumn{1}{c}{\textbf{77.35}} & \multicolumn{1}{c}{\textbf{73.51}} \\

\hline
\multicolumn{12}{p{300pt}}{Rank at r accuracy(\%), mAP(\%) and mINP(\%) are reported. The best results are in bold.}\\
\end{tabular}
}
\label{table2}
\end{table}

\subsubsection{Effectiveness of EMCC}
The participation of EMCC yields a $4.65\%$ and $7.67\%$ improvement of rank-1 on all-search and indoor-search. This proves that improving the clustering algorithm can effectively assign pseudo-labels that are modality- and camera-independent, thus facilitating the learning of modality-camera invariant representations.
\subsubsection{Effectiveness of TSMem}
In experiments 1 to 5, we applied RTMem, which has been shown to be a more effective strategy than the momentum update strategy in \cite{yin2023real}. In experiment 6, the accuracy was improved when we replaced RTMem with our TSMem. The result substantiates that the TSMem keeps the consistency of clustering and the memory updating and improves the generalization of information stored in memory.

\section{Conclusion}

In this letter, we investigate a meaningful and challenging task, unsupervised visible infrared person re-identification (US-VI-ReID), to eliminate the reliance on cross-modal annotations and significantly reduce the cost. In order to introduce cross-modal clustering while avoiding the extreme pursuit of cluster-level association, we propose a novel Extended Cross-Modality United Learning (ECUL) framework, incorporating Extended Modality-Camera Clustering (EMCC) and Two-Step Memory Updating Strategy (TSMem). ECUL is based on the cluster-level and instance-level contrastive losses, naturally integrating intra-modality clustering, inter-modality clustering and inter-modality instance selection, thus effectively learning modality-invariant and camera-invariant representations. Through extensive experiments on SYSU-MM01 and RegDB datasets, we compare ECUL with other state-of-the-art methods. The results show that our method outperforms the prevailing unsupervised methods and is even competitive among certain supervised methods.

\bibliographystyle{IEEEtran} 
\bibliography{references} 

\end{document}